\def\year{2018}\relax
\documentclass[letterpaper]{article} 
\usepackage{aaai18}  
\usepackage{times}  
\usepackage{helvet}  
\usepackage{courier}  
\usepackage{amsmath}
\usepackage{amsfonts}
\usepackage{color}
\usepackage{hyperref}
\usepackage{natbib}
\usepackage{bm}
\usepackage{url}  
\usepackage{graphicx}  
\usepackage{float}
\usepackage{booktabs}
\usepackage{placeins}
\usepackage{rotating}
\restylefloat{table}
\nocopyright
 
\begin{document}
%
\title{Attention-based Mixture Density Recurrent Networks for History-based Recommendation}

\newcommand*{\affaddr}[1]{#1} 
\newcommand*{\affmark}[1][*]{\textsuperscript{#1}}
\newcommand*{\email}[1]{\texttt{#1}}
\newcommand{\alert}[1]{{\color{red} #1}}

\author{
Tian Wang\affmark[1] 
and Kyunghyun Cho\affmark[2]\\
\affaddr{\affmark[1] eBay, New York, USA}\\
\affaddr{\affmark[2] New York University, New York, USA}\\
}

\maketitle
\begin{abstract}
The goal of personalized history-based recommendation is to automatically output a distribution over all the items given a sequence of previous purchases of a user. In this work, we present a novel approach that uses a recurrent network for summarizing the history of purchases, continuous vectors representing items for scalability, and a novel attention-based recurrent mixture density network, which outputs each component in a mixture sequentially, for modelling a multi-modal conditional distribution. We evaluate the proposed approach on two publicly available datasets, MovieLens-20M and RecSys15. The experiments show that the proposed approach, which explicitly models the multi-modal nature of the predictive distribution, is able to improve the performance over various baselines in terms of precision, recall and nDCG. 
\end{abstract}

\section{Introduction}

\noindent 
Recommender systems have become an indispensable part of the e-commerce industry, helping customers to sort out items of interest from large inventories. Among the most popular techniques are matrix factorization (MF) based models~\citep[see, e.g.,][]{hu2008collaborative, koren2009matrix, rendle2009bpr} which decompose a user--item matrix into user and item matrices. Such an approach treats recommendation as a matrix completion/imputation problem, where missing entries in the original matrix are estimated by the dot product between corresponding user and item factors. Despite their popularity in recommender systems, MF-based models have their limitations. First, MF-based models aim at reconstructing user history, instead of predicting future behaviors. The underlying assumption is that user preference is static over time. Second, most of such approaches omit ordering information in a user history. To address these issues, an increasing number of recent works have begun to treat user behaviors as sequences, and predict future events based on history~\citep[see, e.g.,][]{hidasi2015session, tan2016improved, de2017large}. 

Recurrent neural networks (RNNs) are one of the most widely used techniques for sequence modelling \citep[see, e.g.,][]{mikolov2010recurrent,bahdanau2014neural,sutskever2014sequence}. These RNNs have recently been considered for a history-based recommendation system \citep[see, e.g.,][]{hidasi2015session,tan2016improved,wu2017recurrent}. In the implicit feedback scenario, where binary user-item interactions are recorded, the previously mentioned works pose recommendation as a classification task. For a recommender system containing millions of items, classification approach will calculate a score for each user-item pair, therefore leading to a scalability issue in both training and prediction. Besides scalability, in the work by \citet{de2017large}, they demonstrated such an approach fails to recommend relevant items to users on their dataset containing more than 6 million songs.

Instead, \citet{de2017large} formulated the problem as regression rather than classification to address the scalability and performance degradation issues. However, due to the noisy and multi-modal nature of user behavior, building a mapping from history to future is an ill-posed problem, and regression may not be suitable.

In this work, we present an approach to history-based recommendation by generating a conditional distribution over future items vectors. The proposed model uses a recurrent network for summarizing a user history, and an attention-based recurrent mixture density network, which generates each component in a mixture network sequentially, for modelling a multi-modal conditional distribution.

Our model is evaluated on MovieLens-20M \citep{harper2016movielens} and RecSys15~\footnote{http://2015.recsyschallenge.com/} in an implicit feedback setting. Experimental results demonstrate that the proposed model improves recall, precision, and nDCG significantly, compared to various baselines. Comprehensive analysis on model configuration shows that increasing the number of mixture components improves recommendations by better capturing multi-modality in user behavior. 

Our model explores a new direction of recommender systems by conducting density estimation over continuous item representation. It provides an effective way to generate components for a mixture network, which potentially benefits all applications using multiple mixtures. 

\section{Related Work}
\subsection{A History-based Recommender System with a Recurrent Neural Network}

RNNs were first used to model a user history in recommender systems by \citet{hidasi2015session}. In this work, a RNN was used to model previous items and predict the next one in a user sequence. \citet{tan2016improved} improved the recommender system performance on a similar architecture with data augmentation. To better leverage item features, \citet{hidasi2016parallel} introduced a parallel RNN architecture to jointly model user behaviors and item features.  \citet{wu2017recurrent} proposed a new architecture using separate recurrent neural networks to update user and item representations in a temporal fashion.

There are two major differences in the proposed approach from the previously mentioned work. First, our work frames the task of implicit-feedback recommendation as density estimation in a continuous space rather than classification with a discrete output space. Second, unlike most of the earlier works, where the whole systems were trained end-to-end, the proposed model leverages an external algorithm to extract item representation, allowing the system to cope with new items more easily.

More recently, \citet{de2017large} proposed a history-based recommender system with pretrained continuous item representations as a regression problem. In their work, a recurrent neural network read through a user's history, as a sequence of listened songs, and extracted a fixed-length user taste vector, which was later used to predict future songs.

The major difference between the proposed work and the work by \citet{de2017large} is the assumption on the number of modes in the distribution of user behaviors. The proposed model considers the mapping from history to a future behaviour as a probability distribution with multiple modes, unlike their work in which such a distribution is assumed to be unimodal. We do so by using a variant of mixture density network~\citep{bishop1994mixture} to explicitly model user behavior. Their approach can be considered as a special case of the proposed model with a single mixture component.

%
%
%
%
%

\subsection{Continuous Item Representation}

Inspired by the recent advance in word representation learning \citep{mikolov2013efficient}, various methods have been proposed to embed an item in a distributed vector that encodes useful information for recommendation. \citet{barkan2016item2vec} learned item vectors using an approach similar to Word2Vec \citep{mikolov2013efficient} by treating each item as a word without considering the order of items. \citet{liang2016factorization} jointly factorized a user-item matrix and item-item matrix to obtain item vectors. In the work by \citet{liu2017related}, a vector was learned for each pin in Pinterest using a Word2Vec inspired model. 

In this paper, we use external knowledge to extract item representation, instead of training jointly. Such an approach is effective, because it enables the use of any recent advances in representation learning and has a potential to incorporate new items unseen during training.

%
%
%

\section{Model}

\subsection{Recommendation Framework}

In the implicit feedback setting, a user behavior is recorded as a sequence of interacted items, which can be a mixture of various behaviors, including viewing, purchasing, searching and others. For simplicity, we only focus on the viewing behavior in our model. We frame the task of  recommendation as a sequence modelling problem with the goal of predicting the future directly. 

Given a splitting index $t$ and a user behavior sequence $S = \{s_1, s_2, ..., s_L\}$, the sequence can be split into the history $S_{<t} = \{s_1, ..., s_{t-1}\}$ and the future $S_{\geq t} = \{s_t, ..., s_L\}$. A recommender system, parametrized by $w$, aims at modelling the probability of future items conditioned on historical items $P(S_{\geq t} | S_{<t}, w)$. For simplicity, we omit $w$ in our notation and assume that the items in $S_{\geq t}$ are independent
\begin{equation}
P(S_{\geq t} | S_{<t}) = \prod_{i=t}^L P(s_i | S_{<t}).
\end{equation}
This conditional probability  $P(s_i | S_{<t})$ can be approximated by, for instance, $n$-gram conditional probability
\begin{equation} \label{ngram}
P(s_i | S_{<t}) \approx P(s_i| S_{t-1}^{t-(n-1)}),
\end{equation}
where $S_{t-1}^{t-(n-1)} = \{ s_{t-1}, s_{t-2}, ..., s_{t-(n-1)}\}$ are previous $n-1$ viewed items. 

An $n$-gram statistics table records the number of occurrence for each item n-gram in the training corpus. Based on this, the approximated conditional probability can be expressed as 
\begin{equation}
P(s_i| S_{t-1}^{t-(n-1)}) = \frac{c(s_i, s_{t-1}, s_{t-2}, ..., s_{t-(n-1)})}{\sum_j c(s_j, s_{t-1}, s_{t-2}, ..., s_{t-(n-1)})},
\end{equation}
where $c(\cdot)$ is the count in the training corpus. When $n$ equals two, such setting is a similar variant of item-to-item collaborative filtering \citep{linden2003amazon}, where the temporal dependency among items is ignored.

Conditioned on a seed item $s_j$ in a user history, item-to-item collaborative filtering recommends item $\hat{s}$ having the highest co-view probability
\begin{equation}
\hat{s} = \arg\max_s P(s | s_j).
\end{equation}
The statistics table contains the number of occurrence for each item pair $c(s_i, s_j), \text{where } s_i \in S_{\geq t}, s_j \in S_{< t}$.  One can estimate item-to-item conditional probability by
\begin{equation}
P(s_i | s_j) = \frac{c(s_i, s_j)}{\sum_k c(s_k, s_j)} = \frac{c(s_i, s_j)}{c(s_j)}
\end{equation}

With pairwise conditional probability, $P(s_i| S_{t-1}^{t-(n-1)})$ can be approximated using  $P(s_i| s_k)$ by random sampling an item $s_k \in S_{t-1}^{t-(n-1)}$. To stabilize the result, we take the average of the approximated probability
\begin{equation} \label{item-to-item}
P(s_i| S_{t-1}^{t-(n-1)}) \approx \sum_{k = t-1} ^{t-(n-1)}\frac{P(s_i | s_k)}{n-1}.
\end{equation}
A major limitation of such count-based method is data sparsity, 
as a large number of $n$-grams do not occur in the training corpus. 

To address the data sparsity issue in count-based methods, \citet{bengio2003neural} proposed neural language model, in which each word is represented as a continuous vector. In this paper, we take a similar approach by representing each item $s_i$ using a continuous vector $\mathbf{v}_i$. Unlike earlier works using continuous representation as input only, and model its discrete probability distribution as classification, we instead choose to directly model probability density function over continuous item representation $f_p(\mathbf{v}_i | S_{<t})$. $k$ items with highest likelihoods are recommended accordingly. 

In doing so, there are three major technical questions. The first question is how to construct item vector $v_i$. The second one is how to represent a user history $S_{<t}$. Lastly, we must decided how to construct such a probability density function $f_p$. We will answer each of these questions in the following subsections.  

\subsection{Item Representation}

In our model, item embeddings are pretrained and kept fixed during training. 
Although no assumption is posed on item embeddings, the distance between two embeddings should be able to explain certain relationships between two items, such as content similarity and co-purchase likelihood. That is, the closer the distance is the stronger the relationship should be.  

In this paper, we train item embeddings in a way similar to continuous bag-of-words model \citep{mikolov2013efficient} by treating a user sequence of items as a sentence, and each item as a word. Under this setting, the distance between two items in vector space could be explained by their co-occurrence chance in a sequence. The closer the distance is, the higher chance two items have occurring in the same sequence. 

As a result, a valid item embedding matrix $\mathbf{E}$ is generated, where each row is the representation of an item. We denote that for each item $s_i$, its $d_{\text{emb}}$-dimensional vector representation $\mathbf{v}_i$ could be retrieved $\mathbf{E}(s_i) = \mathbf{v}_i \in \mathbb{R}^{d_{\text{emb}}}$.

\subsection{History Representation}

A user's history is recorded as a sequence of items, either viewed or purchased, $S_{<t} = \{s_1, ..., s_{t-1}\}$. After mapping each item to its vector, we can get a sequence of the item vectors $V_{<t} = \{\mathbf{v}_1, ..., \mathbf{v}_{t-1}\}$. In this paper, we experimented with three alternatives to represent user history. 

\paragraph{Continuous Bag-of-Items Representation (CBoI)}

The first proposed method is to simply bag all the items in $S_{<t}$ into a single vector $\mathbf{s} \in [0, 1]^{|\mathbf{E}|}$. Any element of $\mathbf{s}$ corresponding to the item existing in $S_{<t}$ will be assigned the frequency of that item, and otherwise 0. This vector is multiplied from left by item embedding matrix $\mathbf{E}$
\begin{equation}
\mathbf{p} = \mathbf{E}^\top s.
\end{equation}
We call this representation $\mathbf{p}$ a continuous bag-of-items (CBoI). In this approach, the ordering of history items does not affect the representation.

\paragraph{Recurrent Representation (RNN)}

Recurrent neural networks (RNN) have become one of the most popular techniques for modelling a sequence. Long short-term memory units \citep[LSTM,][]{hochreiter1997long} and gated recurrent units \citep[GRU,][]{cho2014learning} are the two most popular variants of RNNs. In this paper, we work with GRUs, which have the following update rule:
\begin{equation}
\label{eq:gru}
\begin{split}
\mathbf{r}_t & = \mathbf{\sigma}  (\mathbf{W}_r \mathbf{x}_t + \mathbf{U}_r \mathbf{h}_{t-1}) \\
\mathbf{u}_t & = \mathbf{\sigma}  (\mathbf{W}_u \mathbf{x}_t + \mathbf{U}_u (\mathbf{r}_t \odot \mathbf{h}_{t-1})) \\
\tilde{\mathbf{h}}_t & = \tanh(\mathbf{W} \mathbf{x}_t + \mathbf{U} (\mathbf{r}_t \odot \mathbf{h}_{t-1})) \\
\mathbf{h}_t & = (1 - \mathbf{u}_t) \odot \mathbf{h}_{t-1} + \mathbf{u}_t \odot \tilde{\mathbf{h}}_t,
\end{split}
\end{equation}
where $\sigma$ is a sigmoid function, $\mathbf{x}_t$ is the input at the $t$-th timestep, and $\odot$ is element-wise multiplication.

After converting each item into a vector representation, the sequence of item vectors is read by a recurrent neural network. We initialize the recurrent hidden state as 0. For each item in the history, we get 
\begin{equation}\label{eq:recurrent_repre}
\mathbf{z}_l = \phi (\mathbf{v}_l, \mathbf{z}_{l-1}),
\end{equation}
for $l = 1, ..., t-1$. $\phi$  is GRU recurrent activation function defined in Eq. (\ref{eq:gru}).

With $Z_{<t} = \{\mathbf{z}_1, ..., \mathbf{z}_{t-1}\}$, recurrent user representation $\mathbf{p}$ is computed by
\begin{equation}
\mathbf{p} = \frac{\sum_{i=1}^{t-1} \mathbf{z}_i}{t-1}.
\end{equation}

\paragraph{Attention-based Representation (RNN-ATT)}

Inspired by the success of attention mechanism in machine translation \citep{bahdanau2014neural}, the proposed method incorporates attention mechanism into recurrent history representation when using with recurrent decoder later described in Eq. (\ref{rnn_decoder}). After $Z_{<t}$ is generated following the same way mentioned in Eq. (\ref{eq:recurrent_repre}), we use a separate bidirectional recurrent neural network to read $V_{<t}$, and generate a sequence of annotated vectors $A_{<t} = \{\mathbf{a}_1, ..., \mathbf{a}_{t-1}\}$. For a mixture vector $\mathbf{m}_l$, attention-based history representation $\mathbf{p}_l$ is calculated as
\begin{equation}\label{attention}
\mathbf{p}_l = \sum_{i=1}^{t-1} \alpha_{l, i} \mathbf{z}_i,
\end{equation}
where the attention weight $\alpha_{l, i}$ is computed by
\begin{equation}\label{eq:att_weight}
\alpha_{l, i} = \frac{\exp (\text{score}(\mathbf{a}_i, \mathbf{h}_{l}))}{\sum_{j=1}^{t-1} \exp (\text{score}(\mathbf{a}_j, \mathbf{h}_{l}))}.
\end{equation}
In Eq. (\ref{eq:att_weight}), $\mathbf{h}_{l}$ is the hidden state of the recurrent neural network in the decoder calculated in Eq. (\ref{rnn_decoder}), and $\text{score}(\mathbf{a}_j, \mathbf{h}_{l})$ function defines the relevance score of the $j$-th item with respect to $\mathbf{h}_{l}$.

\subsection{Mixture Density Network}

A mixture density network \citep[MDN,][]{bishop1994mixture} formulates the likelihood of an item vector $\mathbf{v}_i$ conditioned on a user history $S_{<t}$ (represented by $\mathbf{p}$) as a linear combination of kernel functions
\begin{equation}
f_p(\mathbf{v}_i | \mathbf{p}) = \sum_{j = 1} ^m \alpha_j(\mathbf{p}) \phi_j (\mathbf{v}_i | \mathbf{p}), 
\end{equation}
where $m$ is the number of components used in the mixture. Each kernel is a multivariate Gaussian density function:
\begin{equation} \label{mdn}
\phi_j(\mathbf{v}_i | \mathbf{p}) = \frac{\exp(-\frac{1}{2}(\mathbf{v}_i - \bm{\mu}_j(\mathbf{p}))^T \bm{\Sigma}_j(\mathbf{p})(\mathbf{v}_i - \bm{\mu}_j(\mathbf{p})))}{\sqrt{|2\pi \bm{\Sigma}_j(\mathbf{p})|}}.
\end{equation}
In order to reduce the computation complexity, the covariance matrix $\bm{\Sigma}_j$ is assumed to be diagonal, containing only entries for element-wise variances. 

We propose two methods for generating parameters of the mixture density network using $k$ components.

\paragraph{Feedforward decoder (FF)}

After a user history is encoded into a single user representation $\mathbf{p} \in \mathbb{R}^{d_\text{hidden}}$, the parameters for the $i$-th mixture--$\bm{\mu}_i(\mathbf{p}) \in \mathbb{R}^{d_{\text{emb}}}$, $\text{diag}(\bm{\Sigma}_i(\mathbf{p})) \in \mathbb{R}_{>0}^{d_{\text{emb}}} $, and $\alpha_i(\mathbf{p}) \in \mathbb{R}_{>0}$-- are generated by
\begin{equation}
\begin{split}
\bm{\mu}_i(\mathbf{p}) & = \tanh(\mathbf{W}_{\mu i}\mathbf{p}), \\
\text{diag}(\mathbf{\Sigma}_i(\mathbf{p})) &= \log(\exp (\mathbf{W}_{\Sigma i}\mathbf{p}) + 1), \\
\alpha_i (\mathbf{p}) &= \frac{\exp(\mathbf{W}_{\alpha i} \mathbf{p})}{\sum_j \exp(\mathbf{W}_{\alpha j} \mathbf{p})},
\end{split}
\end{equation}
where $\mathbf{W}_{\mu i} \in \mathbb{R} ^{d_{\text{emb}} \times d_{\text{hidden}}}$, $\mathbf{W}_{\Sigma i} \in \mathbb{R} ^{d_{\text{emb}} \times d_{\text{hidden}}}$, and $\mathbf{W}_{\alpha i} \in \mathbb{R} ^{1 \times d_{\text{hidden}}}$.

\paragraph{Recurrent decoder (RNN)} 

In addition to the feedforward decoder, we propose a recurrent decoder. For a mixture density network with $k$ components, the recurrent decoder iterates $k$ times. In each iteration, an RNN takes the history representation $\mathbf{p}$ as input and generates the parameters of one mixture. Suppose at the $l$-th step, the $l$-th component's parameters are calculated as
\begin{equation} \label{rnn_decoder}
\begin{split}
\mathbf{m}_l & = \phi(\mathbf{p}, \mathbf{m}_{l-1}), \\
\bm{\mu}_l& = \tanh(\mathbf{W_{\mu}} \mathbf{m}_l), \\
\text{diag}(\bm{\Sigma}_l) & = \log(\exp (\mathbf{W}_{\Sigma} \mathbf{m}_l) + 1),
\end{split}
\end{equation}
where $\phi$ is a recurrent activation function, $\mathbf{W}_{\mu} \in \mathbb{R} ^{d_{\text{emb}} \times d_{\text{hidden}}}$ and $\mathbf{W}_{\Sigma} \in \mathbb{R} ^{d_{\text{emb}} \times d_{\text{hidden}}}$ are shared among all mixtures.

After all $\{\mathbf{m}_i\}_{i=1}^n$ are generated, the mixture weight $\alpha_l$ is calculated by
\begin{equation}
\alpha_l = \frac{\exp(\mathbf{W}_{\alpha} \mathbf{m}_l)}{\sum_j \exp(\mathbf{W}_{\alpha} \mathbf{m}_j)},
\end{equation}
where $\mathbf{W}_{\alpha} \in  \mathbb{R} ^{1 \times d_{\text{hidden}}}$.
\begin{figure}[t]
\centering
	\includegraphics[width=0.65\linewidth]{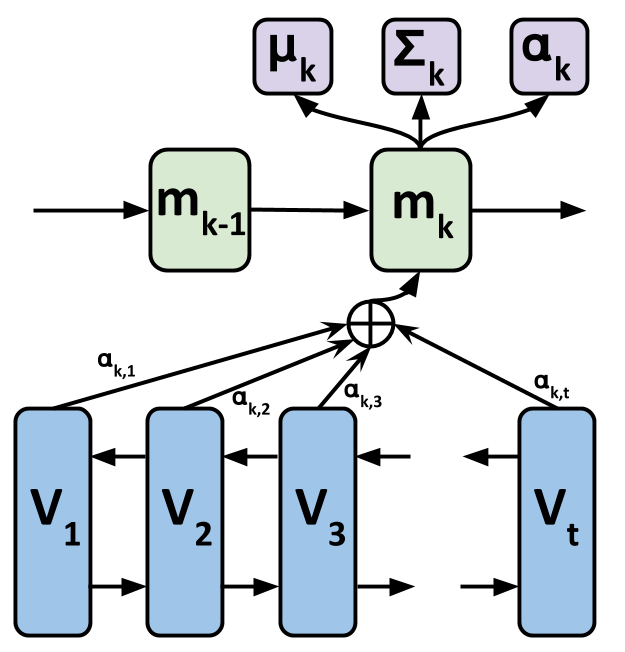}
    \caption{Architecture of recurrent decoder with attention-based history representation}\label{fig:attention}
\end{figure}

Alternatively with the attention-based history representation described in Eq. (\ref{attention}), $\mathbf{p}$ is replaced by $\mathbf{p}_l$ at the $l$-th iteration in Eq. (\ref{rnn_decoder}). At the $l$-th step, for annotated vectors $A_{<t} = \{\mathbf{a}_1, ..., \mathbf{a}_{t-1}\}$, attention weight $\alpha_{l,i}$ is computed by
\begin{equation}
\alpha_{l, i} = \frac{\exp( \text{score}(\mathbf{a}_i, \mathbf{m}_{l-1}))}{\sum_{j=1}^{t-1} \exp (\text{score}(\mathbf{a}_j, \mathbf{m}_{l-1}))}.
\end{equation}
Attention-based recurrent representation $\mathbf{p}_l$ is computed by
\begin{equation}
\mathbf{p}_l = \sum_{i=1}^{t-1} \alpha_{l,i}\mathbf{z}_i,
\end{equation}
where $\mathbf{z}_i$ is the output from Recurrent Representation calculated in Eq. (\ref{eq:recurrent_repre}). The architecture for recurrent decoder with the attention-based encoder is illustrated in Fig.~\ref{fig:attention}. The attention mechanism allows a model to automatically search for items in the user history relevant to each mixture component. 

\section{Experimental Settings}

\subsection{Models}

There are multiple configurations of the proposed methods. First, there are three ways to represent a user history: (1) continuous Bag-of-Items (CBoI), (2) recurrent representation (RNN), and (3) attention-based representation (RNN-ATT). Then, there are two ways to generate mixture parameters: (1) Feedforward decoder (FF) and (2) Recurrent decoder (RNN). We denote all models evaluated in our experiments by
\begin{enumerate}
	\item CBoI-FF-$n$
	\item RNN-FF-$n$
    \item RNN-RNN-$n$
    \item RNN-ATT-RNN-$n$,
\end{enumerate}
where $n$ denotes the number of mixture components in a mixture density network. We test four $n$'s: 1, 2, 4, and 8. 

Note that, when $n$ is equal to 1, a mixture model can only output a unimodal Gaussian distribution. This is similar to the work by \citet{de2017large}, where regression can be viewed as an unimodal Gaussian with an identity covariance matrix. 

We consider following baselines:
\begin{description}

\item[Recently Viewed Items (RVI)] recommends items a user has viewed in the history, ranked by the recency. Although this technique is not a collaborative filtering method, it is widely used as a personalized recommendation module in production systems. In a previous work by \citet{song2016multi}, a similar approach (\textit{Prev-day Click}) was adopted as a baseline method, and outperformed all MF-base model in their experiment. 

\item[Item-to-Item Collaborative Filtering (Item-CF)] uses a single item as a seed instead of using a whole user history, as described in Eq. (\ref{item-to-item}). Recommended items are ranked by an estimated conditional probability.

\item[Implicit Matrix Factorization (IMF)] is implemented according to \citet{hu2008collaborative} by using \textit{implicit} package\footnote{https://github.com/benfred/implicit}. The model is fit using history and future sequences in a training set, and history sequences in validation and testing sets. 

\end{description}

All mixture density network models uses 256 as $d_{\text{hidden}}$, and are trained using Adam~\citep{kingma2014adam} to maximize the log-likelihood defined as
\begin{equation}
\mathcal{L}(\theta) = \frac{1}{K}\sum_{k=1}^K \frac{\sum_{i=t}^{L^k} \log(f_p(s^k_i | S^k_{<t}, \theta)}{L^k -t + 1},
\end{equation}
where $K$ is the number of user sequences in the training set, and $L^k$ is the length of the $k$-th sequence.

In all RNN-based models, a one-layer GRU with 256 hidden units is used. We early-stop training based on F1@20 on a validation set, and report metric on a test set using the best model according to the validation performance.

For implicit matrix factorization, we perform grid search over the number of factors on a validation set, and report the metric using the best model on a test set.

Item embeddings are trained using the continuous bag-of-words model from \textit{FastText} package \citep{bojanowski2016enriching}, with the item embedding dimension set to 100 and windows size to 5. All sequences in the training set are used for embedding learnings. After training, each item vector is normalized by $l_2$ norm. 

\subsection{Datasets}

We evaluated our model on two publicly available datasets. 

\subsubsection{MovieLens-20M}

MovieLens-20M~\citep{harper2016movielens} is a classic explicit-feedback collaborative filtering dataset for movie recommendation, in which (user, movie, rating, timestamp) tuples are recorded. We transform MovieLens-20M into an implicit-feedback dataset by only taking records having ratings greater or equal to 4 as positive observations. User behavior sequences are sorted by time, and those containing more than 15 implicit positive observations are included. Each last viewed 15 movies by each user are split into 10 and 5, as history and future respectively. As the nature of this dataset, there is no duplicate items in the user sequence. After preprocessing, 75,962 sequences are kept. 80\%, 10\%, and 10\% of sequences are randomly split into training, validation, and test sets, respectively. A movie vocabulary is built using the training set, containing 16,253 unique movies. 

\subsubsection{RecSys15}

RecSys15~\footnote{http://2015.recsyschallenge.com/} is an implicit feedback dataset, containing click and purchase events from an online e-commerce website. We only work with the training file in the original dataset, and keep the click events with timestamps. We filter sequences of length less than 15, and use final 2 clicks as future, and the first 13 clicks as history. We do not filter out duplicate items, and as a result the same item could appear in both history and target parts. After preprocessing, we are left with 168,202 sequences. 80\%, 10\%, and 10\% of sequences are randomly split into training, validation, and test sets, respectively. An item vocabulary is built only using items in the training set, leaving us with 32,117 unique items. 

\begin{figure*}[t!]
	\small
    \centering
    \begin{minipage}{0.95\textwidth}
        \centering
        \includegraphics[width=0.6\columnwidth]
        {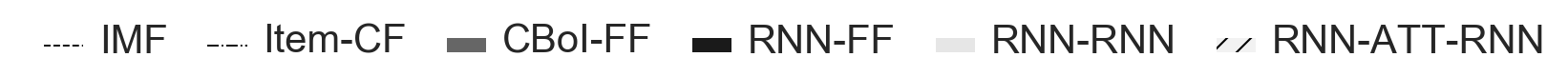}
    \end{minipage}
    \begin{minipage}{0.32\textwidth}
        \centering
        \includegraphics[width=\columnwidth]	{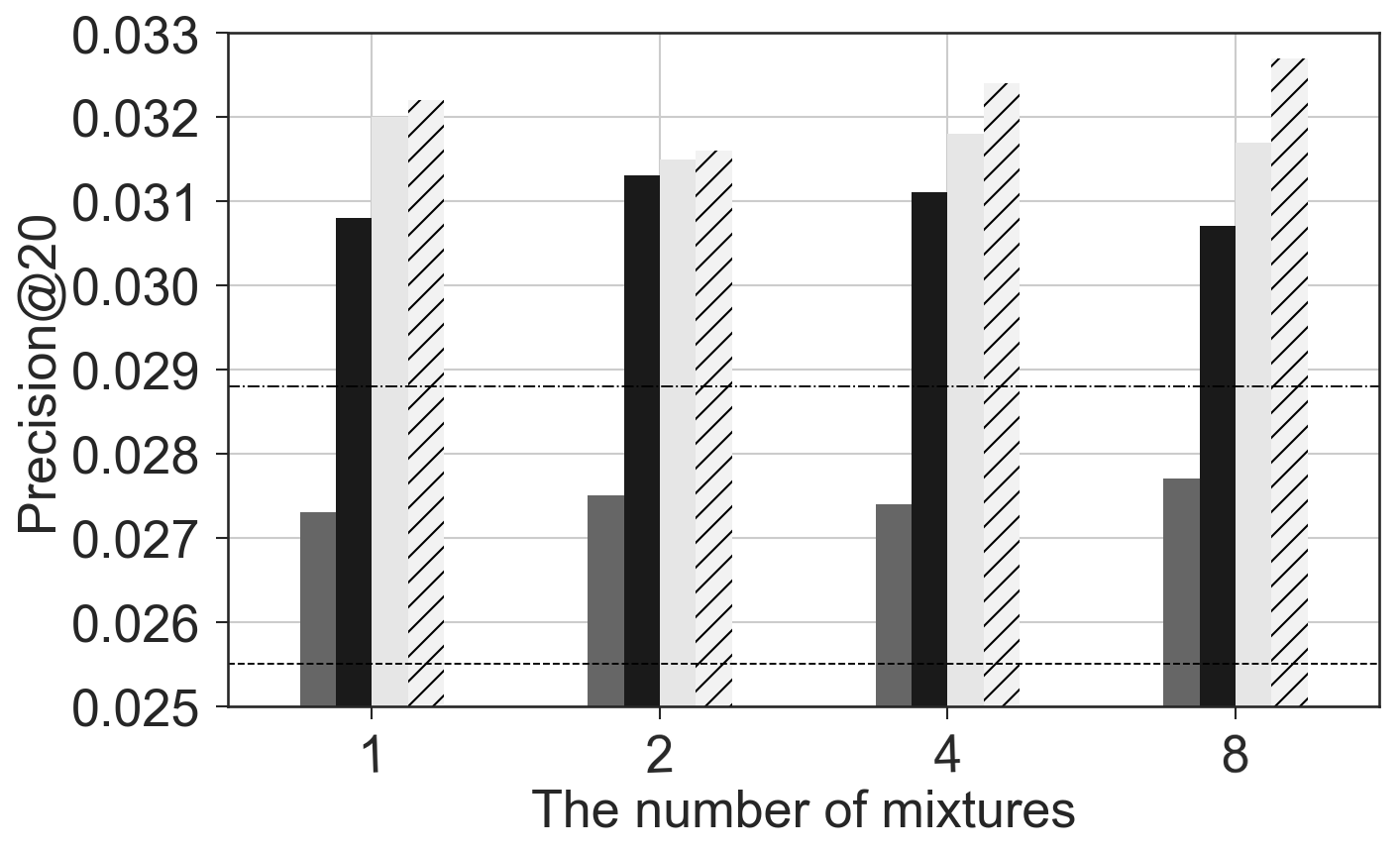}
    \end{minipage}
    \hfill
    \begin{minipage}{0.32\textwidth}
        \centering
        \includegraphics[width=\columnwidth]{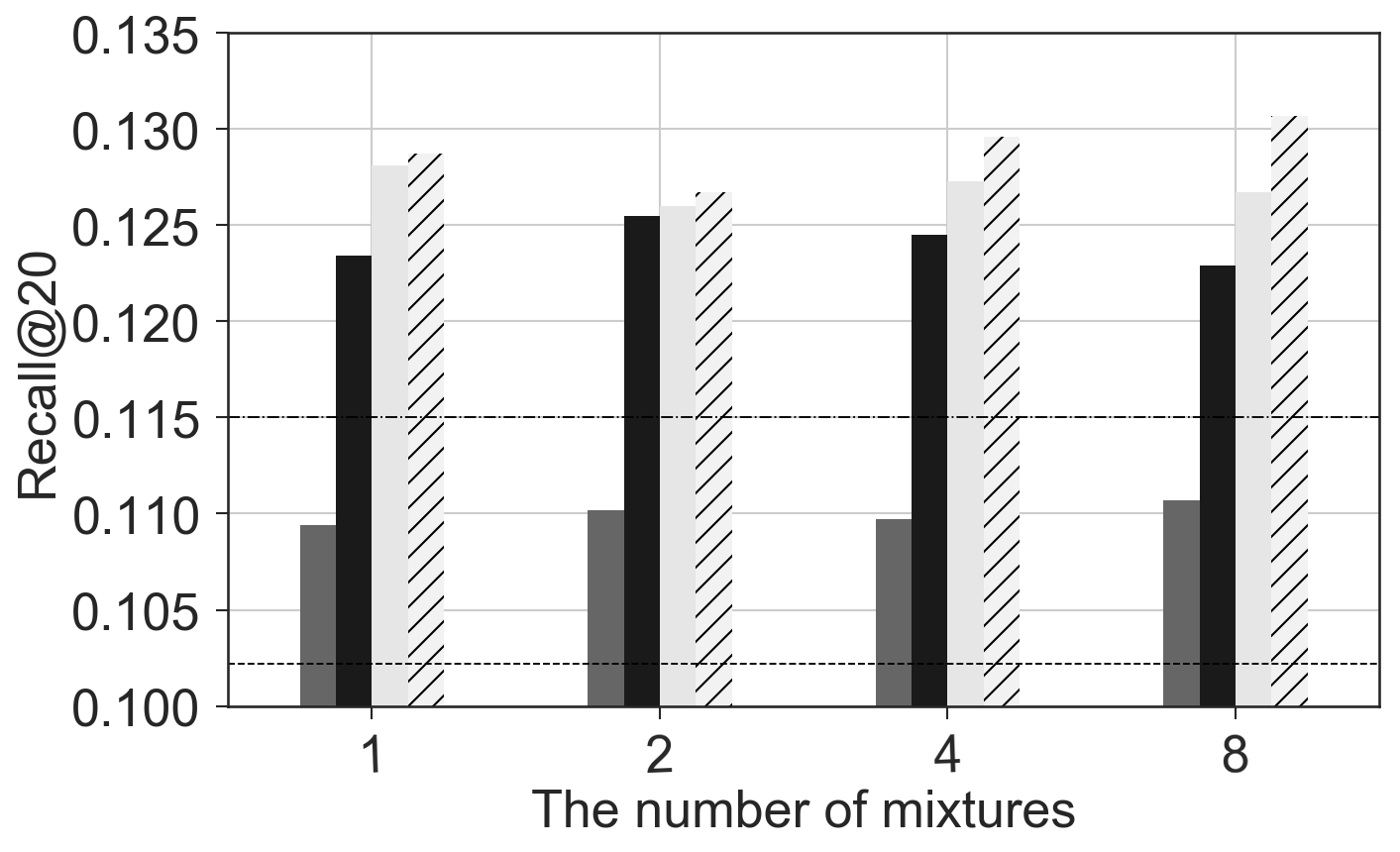}
    \end{minipage}
    \hfill
    \begin{minipage}{0.32\textwidth}
        \centering
        \includegraphics[width=\columnwidth]{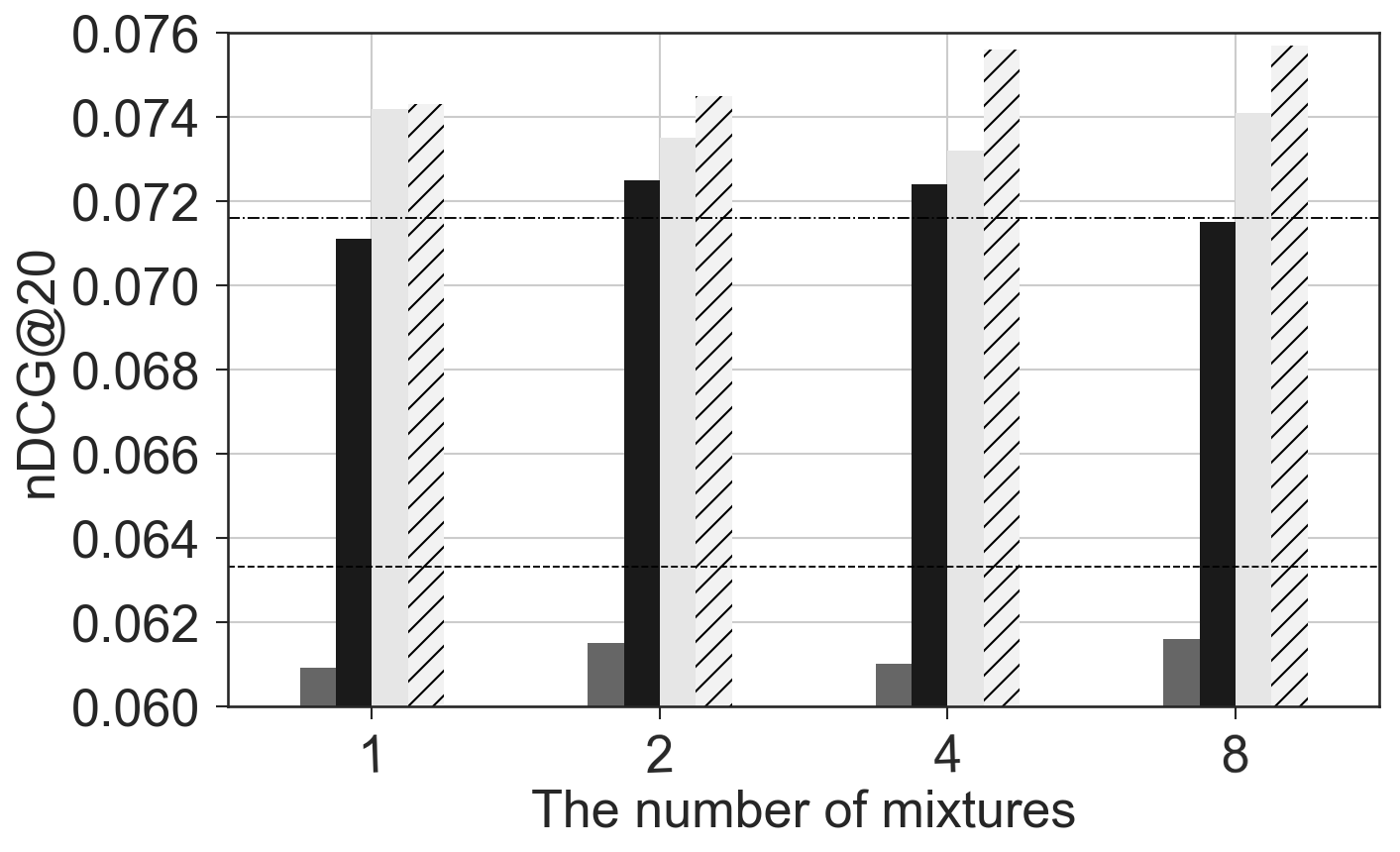}
    \end{minipage}

    \begin{minipage}{0.95\textwidth}
        \centering
        (a) MovieLens-20M
    \end{minipage}
        \begin{minipage}{0.95\textwidth}
        \centering
        \includegraphics[width=0.6\columnwidth]
        {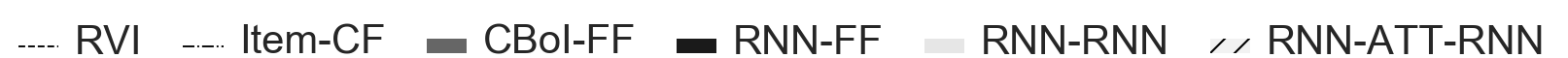}
    \end{minipage}
    \begin{minipage}{0.32\textwidth}
        \centering
        \includegraphics[width=\columnwidth]	{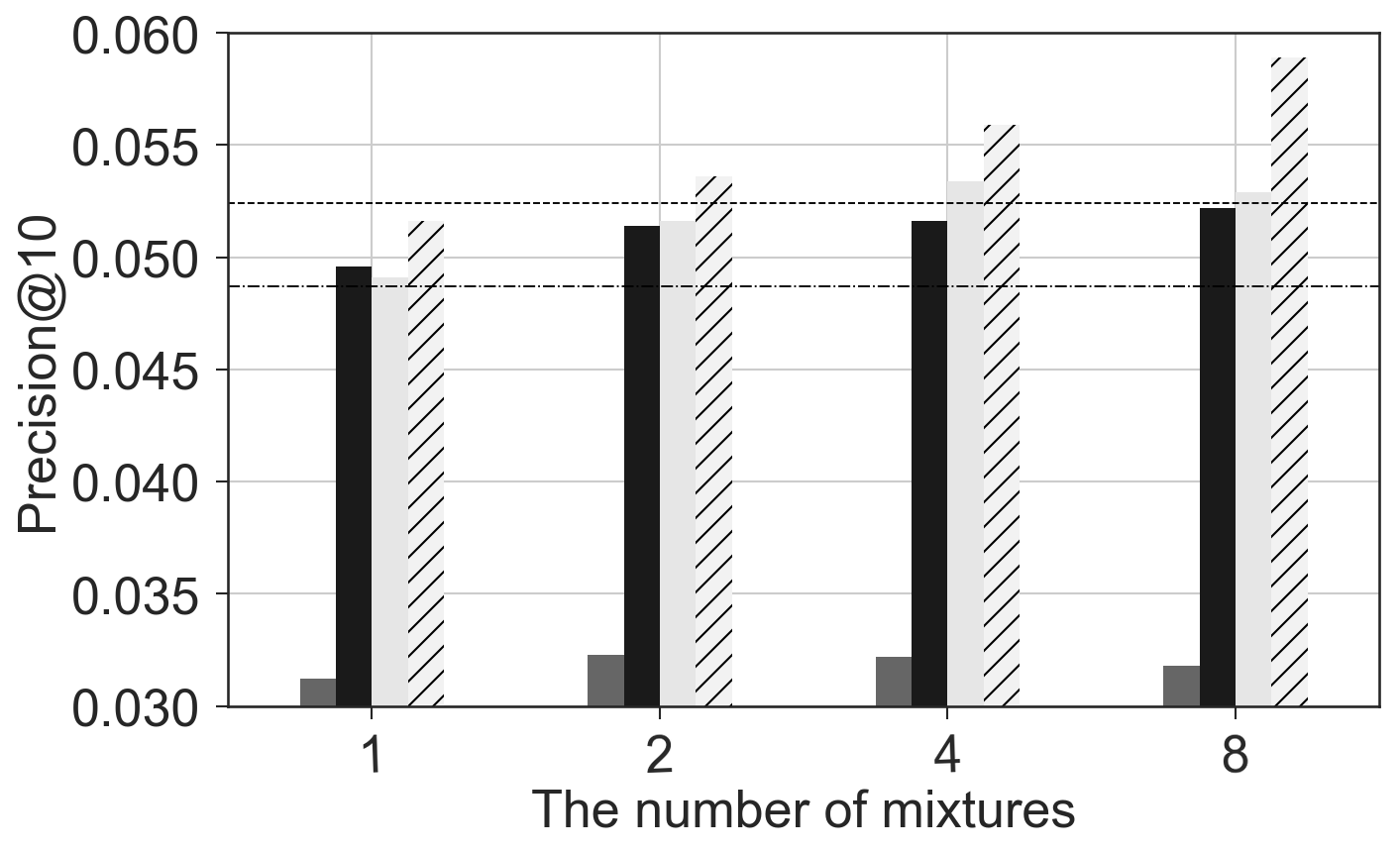}
    \end{minipage}
    \hfill
    \begin{minipage}{0.32\textwidth}
        \centering
        \includegraphics[width=\columnwidth]{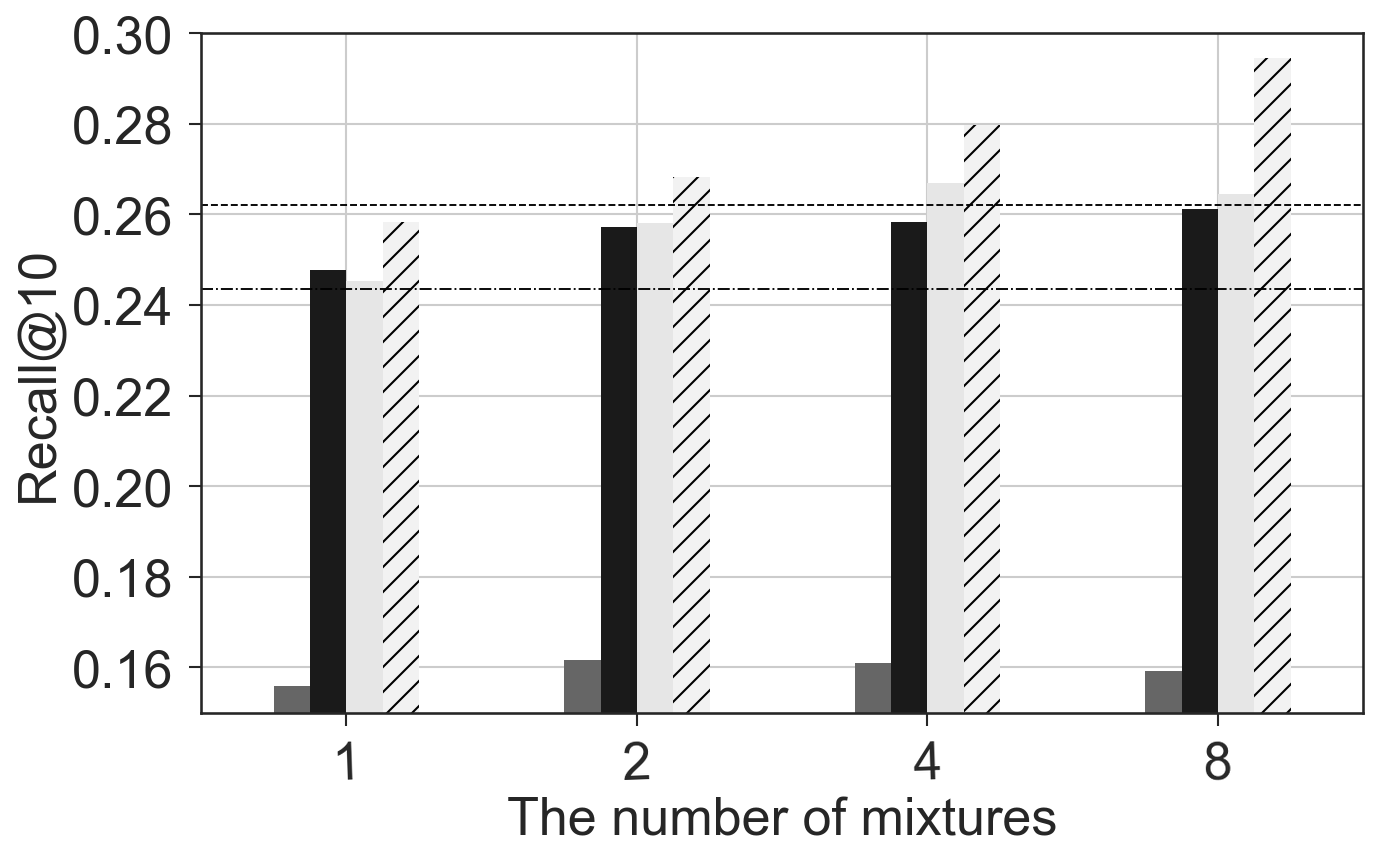}
    \end{minipage}
    \hfill
    \begin{minipage}{0.32\textwidth}
        \centering
        \includegraphics[width=\columnwidth]{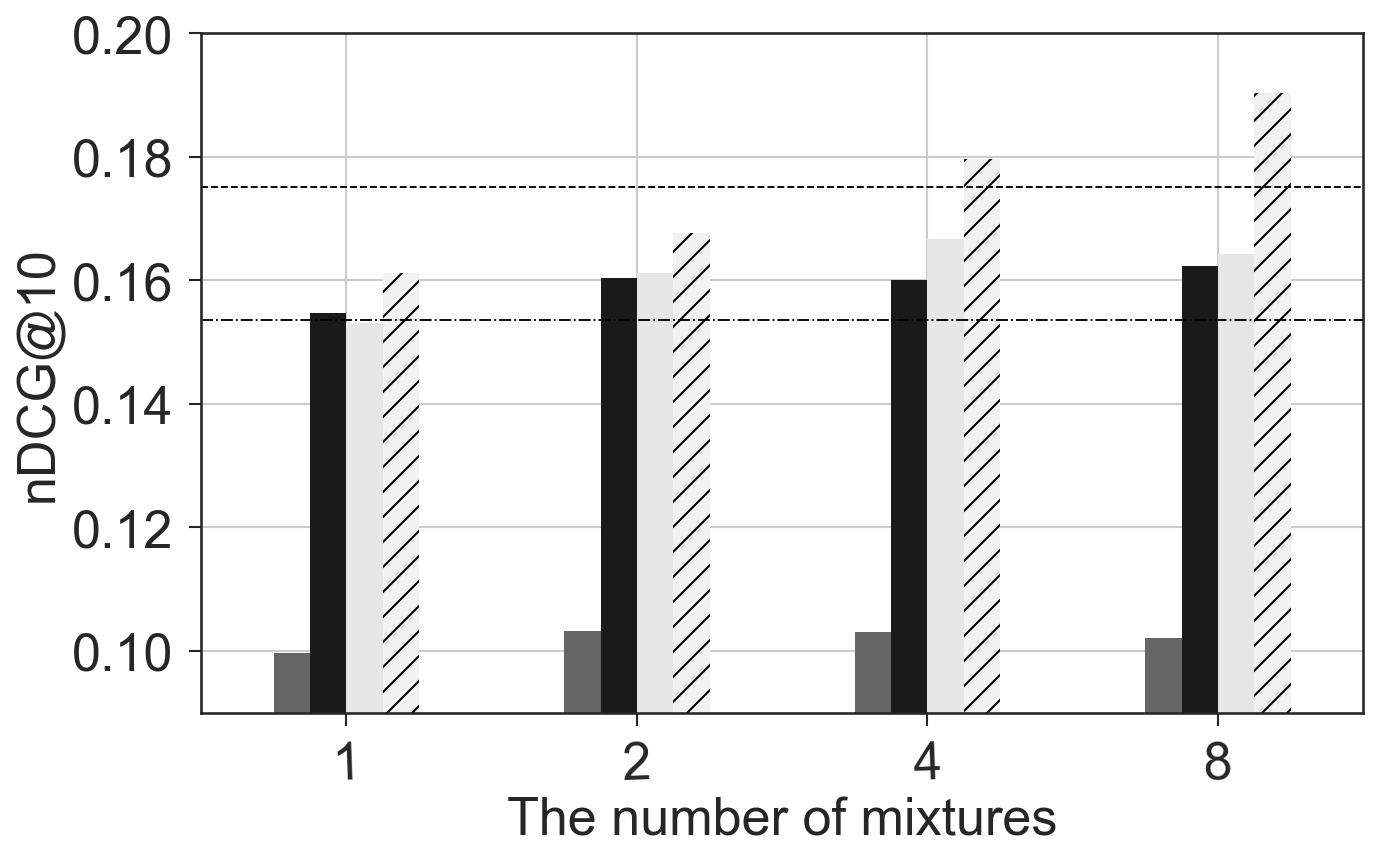}
    \end{minipage}

    \begin{minipage}{0.95\textwidth}
        \centering
        (b) RecSys15 \footnotemark
    \end{minipage}
    \caption{Precision, Recall, and nDCG with varying number of mixture components on (a) MovieLens-20M and (b) RecSys15.}
    \label{fig:metric}
\end{figure*}

\subsection{Metric}

There are various metrics that could be used to evaluate the performance of a recommender system. In this paper, we use precision, recall, and nDCG. Higher values indicate better performance under these metrics. 

We denote top-k recommended items by $R_k = \{r_1, r_2, ...,r_k\}$, where the items are ranked by the recommendation system; and we denote target items by $T = \{t_1, t_2, ..., t_n\}$.
\begin{description}
\item[Precision@k] calculates the fraction of top-k recommended items which are overlapping with target items.
\begin{equation}
\text{Precision@k} = \frac{|R_k \cap T|}{k}
\end{equation}

\item[Recall@k] calculates the fraction of target items which are overlapping with top-k recommended items.

\begin{equation}
\text{Recall@k} = \frac{|R_k \cap T|}{| T|}
\end{equation}
%

\item[nDCG@k] computes the quality of ranking, by comparing the recommendation DCG with the optimal DCG~\citep{jarvelin2002cumulated}. In implicit feedback datasets, relevance scores for items in the target set are assigned to 1. DCG@k is calculated as
\begin{equation}
\text{DCG@k} = \sum_{i= 1} ^{|R_k|}\frac{\mathbb{I}(r_i \in T)}{\log_2(i + 1)}.
\end{equation}
The optimal DCG is calculated as
\begin{equation}
\text{DCG}_\text{optimal} = \sum_{i  = 1} ^{|T|}\frac{1}{\log_2(i + 1)}.
\end{equation}nDCG@k is calculated as
\begin{equation}
\text{nDCG@k} = \frac{\text{DCG@k}}{\text{DCG}_\text{optimal}}.
\end{equation}
\end{description}

\section{Result}

Table~\ref{result} summarizes the results of our experiments. As MovieLens has no duplicate items in a sequence, RVI is not used on that dataset. From the result on MovieLens, we first observe that the proposed RNN-ATT-RNN-{4,8} model consistently outperformed the other methods in all the metrics by large margins. Second, we see that CBoI-FF does not work well regardless of the number of components used, while the performance is substantially improved with the recurrent encoder. Third, comparing between the two baseline models, Item-CF outperforms IMF by a good margin across all metrics. 

On RecSys15, besides the similar trends we see from MovieLens, there are several new observations. First, RVI outperforms all the models except for the RNN-ATT-RNN-\{2,4,8\} on Precision@10 and Recall@10. This result is in line with \citet{song2016multi}. They also observed the competitive performance from using previous day's clicks on news recommendation. Secondly, IMF is the worst performing model on this dataset. We conjecture that this is because that IMF only recommends items a user has not interacted with, while in clicking streams like RecSys15, items in the history are likely to reappear in the future.

To better understand the effect of mixture components on various model architecture, we group by the number of component used across various models and visualize the result in Fig.~\ref{fig:metric}. We observe that RNN-ATT-RNN-$n$ achieves most visible improvement as the number $n$ of mixture components increases. We also notice that for all MDN-based architectures, using two mixtures always achieves better result compared with using one mixture. However, unless the attention mechanism is used we see diminishing improvements with more components.

The experiments have revealed that it is clearly beneficial to capture the multimodal nature of prediction in a recommender system. This is however only possible with the right choice of user representation and right mechanism for generating mixture parameters. In these experiments, our novel approach, the attention-based recurrent history representation combined with the recurrent decoder, was found to be the best choice in both datasets. We have further learned that the user preference is not static across time, and that it is beneficial to model the user history as a sequence rather than a bag.

\footnotetext{Implicit Matrix Factorization resulted in the score of 0.0176, 0.0878, and 0.0402 respectively for precision@10, recall@10, and nDCG@10, and is not shown in here}

\section{Conclusion \& Future Work}

In this paper, we proposed a method to construct a recommender system by generating the probability density function over future item vectors. The proposed model combines recurrent user history representation with a mixture density network, where a novel attention-based recurrent mixture density has been proposed to output each mixture component sequentially. The experiments on two publicly available datasets, MovieLens-20M and RecSys 15, have demonstrated significant improvement in recall, precision, and nDCG compared against various baselines, validating the advantage of modelling the multimodal nature of the predictive distribution in a recommendation system.


\begin{table*}[htb]
	\centering
    \begin{minipage}{0.16\textwidth}
    \resizebox{\linewidth}{!}{
    	\begin{tabular}{@{}c|c@{}}
        	\toprule
            Model   & P@10 \\
            \midrule
            CBoI-FF-4    &0.0283     \\
			CBoI-FF-1   & 0.0286    \\
            CBoI-FF-8    & 0.0286     \\
            CBoI-FF-2    & 0.0289  \\
            IMF     &  0.0301     \\
            Item-CF & 0.0337    \\
            RNN-FF-1     & 0.0343    \\
            RNN-FF-2     & 0.0348    \\
            RNN-FF-4     & 0.0350     \\
            RNN-RNN-4      & 0.0350   \\
            RNN-FF-8     & 0.0351   \\
            RNN-RNN-1      & 0.0352    \\
            RNN-RNN-2      & 0.0352     \\
            RNN-ATT-RNN-1     & 0.0356    \\
            RNN-ATT-RNN-2     & 0.0356     \\
            RNN-RNN-8      & 0.0358     \\
            RNN-ATT-RNN-8     & 0.0363    \\            
            RNN-ATT-RNN-4     & 0.0365    \\
            \midrule
         \end{tabular}
        }
     \end{minipage}
     \begin{minipage}{0.16\textwidth}
      \resizebox{\linewidth}{!}{
    	\begin{tabular}{@{}c|c@{}}
        	\toprule
            Model   & P@20 \\
            \midrule
            IMF           & 0.0255       \\
            CBoI-FF-1    & 0.0273     \\
            CBoI-FF-4    & 0.0274     \\
            CBoI-FF-2    & 0.0275     \\
            CBoI-FF-8    & 0.0277      \\
            Item-CF & 0.0288   \\
            RNN-FF-8    & 0.0307     \\
            RNN-FF-1    & 0.0308     \\
            RNN-FF-4    & 0.0311    \\
            RNN-FF-2    & 0.0313       \\
            RNN-RNN-2     & 0.0315     \\
            RNN-ATT-RNN-2     & 0.0316     \\ 
            RNN-RNN-8     & 0.0317    \\
            RNN-RNN-4     & 0.0318      \\
            RNN-RNN-1     & 0.0320       \\
            RNN-ATT-RNN-1     & 0.0322      \\           
            RNN-ATT-RNN-4    & 0.0324     \\
            RNN-ATT-RNN-8    & 0.0327  \\
            \midrule
         \end{tabular}
        }
     \end{minipage}
     \begin{minipage}{0.16\textwidth}
     \resizebox{\linewidth}{!}{
    	\begin{tabular}{@{}c|c@{}}
        	\toprule
            Model   & R@10 \\
            \midrule
            CBoI-FF-4    &  0.0567     \\
            CBoI-FF-1    & 0.0573     \\
            CBoI-FF-2    & 0.0573     \\
 	        CBoI-FF-8    & 0.0578     \\
            IMF           & 0.0603       \\
            Item-CF & 0.0674   \\
		    RNN-FF-1    & 0.0686     \\
            RNN-FF-2    & 0.0695       \\
            RNN-FF-4    & 0.0700     \\
            RNN-RNN-4     &  0.0701      \\
            RNN-FF-8    & 0.0702     \\
            RNN-RNN-1     & 0.0705       \\
            RNN-RNN-2     &  0.0705     \\
            RNN-ATT-RNN-1     &  0.0712      \\
            RNN-ATT-RNN-2     & 0.0712    \\ 
            RNN-RNN-8     & 0.0717    \\
            RNN-ATT-RNN-8    & 0.0727  \\
            RNN-ATT-RNN-4    & 0.0731    \\
            \midrule
         \end{tabular}
        }
     \end{minipage}
     \begin{minipage}{0.16\textwidth}
      \resizebox{\linewidth}{!}{
    	\begin{tabular}{@{}c|c@{}}
        	\toprule
            Model   & R@20 \\
            \midrule
            IMF           & 0.1022        \\
            CBoI-FF-1    & 0.1094     \\
            CBoI-FF-4    &  0.1097     \\
            CBoI-FF-2    &  0.1102     \\
            CBoI-FF-8    & 0.1107     \\
            Item-CF & 0.1150   \\
            RNN-FF-8    & 0.1229     \\
		    RNN-FF-1    & 0.1234     \\
            RNN-FF-4    & 0.1245     \\
            RNN-FF-2    & 0.1255       \\
            RNN-RNN-2     &  0.1260     \\
            RNN-RNN-8     & 0.1267     \\
            RNN-ATT-RNN-2     & 0.1267     \\ 
            RNN-RNN-4     &  0.1273      \\
            RNN-RNN-1     & 0.1281       \\
            RNN-ATT-RNN-1     &  0.1287      \\
            RNN-ATT-RNN-4    & 0.1296     \\
            RNN-ATT-RNN-8    & 0.1307\\
            \midrule
         \end{tabular}
         }
     \end{minipage}
     \begin{minipage}{0.16\textwidth}
      \resizebox{\linewidth}{!}{
    	\begin{tabular}{@{}c|c@{}}
        	\toprule
            Model   & \tiny{nDCG}@20 \\
            \midrule
            CBoI-FF-1    & 0.0422    \\
            CBoI-FF-8    & 0.0424     \\
            CBoI-FF-4    &  0.0426     \\
            CBoI-FF-2    &  0.0426     \\
             IMF           & 0.0492        \\
		    RNN-FF-1    & 0.0514     \\
            RNN-FF-2    & 0.0524       \\
            RNN-FF-8    & 0.0525     \\
            RNN-RNN-4     &  0.0526      \\
            RNN-FF-4    & 0.0528     \\
            RNN-RNN-1     & 0.0534       \\
            RNN-RNN-2     &  0.0535     \\ 
            RNN-ATT-RNN-1     &  0.0536      \\
            RNN-RNN-8     & 0.0542     \\
            Item-CF & 0.0544   \\
            RNN-ATT-RNN-2     & 0.0544     \\
            RNN-ATT-RNN-8    & 0.0548\\
            RNN-ATT-RNN-4    & 0.0552     \\
            \midrule
         \end{tabular}
         }
     \end{minipage}
     \begin{minipage}{0.16\textwidth}
      \resizebox{\linewidth}{!}{
    	\begin{tabular}{@{}c|c@{}}
        	\toprule
            Model   & \tiny{nDCG}@20 \\
            \midrule
            CBoI-FF-1    & 0.0609   \\
            CBoI-FF-4    &  0.0610  \\
            CBoI-FF-2    &  0.0615    \\
            CBoI-FF-8    & 0.0616      \\
            IMF           & 0.0633         \\
		    RNN-FF-1    & 0.0711    \\
            RNN-FF-8    & 0.0715     \\
            Item-CF & 0.0716    \\
            RNN-FF-4    & 0.0724   \\
            RNN-FF-2    & 0.0725      \\
            RNN-RNN-4     &  0.0732     \\
            RNN-RNN-2     &  0.0735  \\ 
            RNN-RNN-8     & 0.0741    \\
            RNN-RNN-1     & 0.0742     \\
            RNN-ATT-RNN-1     &  0.0743     \\
            RNN-ATT-RNN-2     & 0.0745     \\
            RNN-ATT-RNN-4    & 0.0756   \\
            RNN-ATT-RNN-8    & 0.0757\\
            \midrule
         \end{tabular}
         }
	 \end{minipage}
     
     \begin{minipage}{0.9\textwidth}
        \centering
        (a) MovieLens-20M
    \end{minipage}
     
     \begin{minipage}{0.16\textwidth}
    \resizebox{\linewidth}{!}{
    	\begin{tabular}{@{}c|c@{}}
        	\toprule
            Model   & P@10 \\
            \midrule
            IMF     &  0.0176     \\
			CBoI-FF-1   & 0.0312    \\
            CBoI-FF-8    & 0.0318     \\
            CBoI-FF-4    &0.0322      \\
            CBoI-FF-2    & 0.0323  \\
            Item-CF & 0.0487    \\
            RNN-RNN-1      & 0.0491    \\
            RNN-FF-1     & 0.0496    \\
            RNN-FF-2     & 0.0514    \\
            RNN-ATT-RNN-1     & 0.0516    \\
            RNN-RNN-2      & 0.0516     \\
            RNN-FF-4     & 0.0516     \\
            RNN-FF-8     &0.0522   \\
            RVI          & 0.0524  \\
            RNN-RNN-8      & 0.0529     \\
            RNN-RNN-4      & 0.0534   \\
            RNN-ATT-RNN-2     & 0.0536     \\           
            RNN-ATT-RNN-4     & 0.0559    \\
            RNN-ATT-RNN-8     & 0.0589    \\ 
            \midrule
         \end{tabular}
        }
     \end{minipage}
     \begin{minipage}{0.16\textwidth}
      \resizebox{\linewidth}{!}{
    	\begin{tabular}{@{}c|c@{}}
        	\toprule
            Model   & P@20 \\
            \midrule
            IMF           & 0.0127       \\
            CBoI-FF-1    & 0.0215     \\
            CBoI-FF-8    & 0.0215      \\
            CBoI-FF-4    & 0.0217     \\
            CBoI-FF-2    & 0.0217     \\
            RVI           & 0.0273  \\
            RNN-FF-1    & 0.0320     \\
            RNN-FF-2    & 0.0320      \\
            RNN-RNN-1     & 0.0321      \\ 
            RNN-RNN-2     & 0.0324   \\
            RNN-FF-4    & 0.0324     \\
            RNN-FF-8    & 0.0327     \\
            RNN-ATT-RNN-1     & 0.0331      \\ 
            RNN-RNN-8     &  0.0331       \\
            Item-CF & 0.0334   \\
            RNN-RNN-4     & 0.0334     \\
            RNN-ATT-RNN-2     &  0.0336       \\          
            RNN-ATT-RNN-4    & 0.0344     \\
            RNN-ATT-RNN-8    & 0.0358  \\
            \midrule
         \end{tabular}
        }
     \end{minipage}
     \begin{minipage}{0.16\textwidth}
     \resizebox{\linewidth}{!}{
    	\begin{tabular}{@{}c|c@{}}
        	\toprule
            Model   & R@10 \\
            \midrule
            IMF           & 0.0878     \\
            CBoI-FF-1    & 0.1559    \\
            CBoI-FF-8    & 0.1592     \\
            CBoI-FF-4    &  0.1610    \\
 	        CBoI-FF-2    & 0.1616     \\
            Item-CF & 0.2435   \\
            RNN-RNN-1     & 0.2453       \\
		    RNN-FF-1    & 0.2478     \\
            RNN-FF-2    & 0.2571       \\
            RNN-RNN-2     &  0.2581     \\
            RNN-ATT-RNN-1     &  0.2582      \\
            RNN-FF-4    & 0.2582     \\
            RNN-FF-8    & 0.2612     \\
            RVI          & 0.2621  \\
            RNN-RNN-8     & 0.2645    \\
            RNN-RNN-4     &  0.2668      \\
            RNN-ATT-RNN-2     & 0.2682    \\ 
            RNN-ATT-RNN-4    & 0.2797    \\
            RNN-ATT-RNN-8    & 0.2944  \\
            \midrule
         \end{tabular}
        }
     \end{minipage}
     \begin{minipage}{0.16\textwidth}
      \resizebox{\linewidth}{!}{
    	\begin{tabular}{@{}c|c@{}}
        	\toprule
            Model   & R@20 \\
            \midrule
            IMF           & 0.1267     \\
            CBoI-FF-8    & 0.2146     \\
            CBoI-FF-1    & 0.2148    \\
            CBoI-FF-4    &  0.2166    \\
 	        CBoI-FF-2    & 0.2171     \\
            RVI          & 0.2728  \\
            RNN-FF-2    & 0.3198       \\
		    RNN-FF-1    & 0.3201     \\
            RNN-RNN-1     & 0.3212       \\
            RNN-RNN-2     &  0.3235     \\
            RNN-FF-4    & 0.3237     \\
            RNN-FF-8    & 0.3269     \\
            RNN-RNN-8     & 0.3306    \\
            RNN-ATT-RNN-1     &  0.3313       \\
            RNN-RNN-4     &  0.3337     \\
            Item-CF & 0.3342   \\
            RNN-ATT-RNN-2     & 0.3362    \\ 
            RNN-ATT-RNN-4    & 0.3439   \\
            RNN-ATT-RNN-8    & 0.3578 \\
            \midrule
         \end{tabular}
         }
     \end{minipage}
     \begin{minipage}{0.16\textwidth}
      \resizebox{\linewidth}{!}{
    	\begin{tabular}{@{}c|c@{}}
        	\toprule
            Model   & \tiny{nDCG}@10 \\
            \midrule
            IMF           &  0.0402     \\
            CBoI-FF-1    & 0.0996    \\
            CBoI-FF-8    & 0.1020     \\
            CBoI-FF-4    &  0.1031    \\
 	        CBoI-FF-2    & 0.1032      \\
            RNN-RNN-1     & 0.1531     \\
            Item-CF & 0.1536   \\
		    RNN-FF-1    & 0.1547   \\
            RNN-FF-4    & 0.1601     \\
            RNN-FF-2    & 0.1603        \\
            RNN-ATT-RNN-1     &  0.1612       \\
            RNN-RNN-2     &  0.1612     \\
            RNN-FF-8    & 0.1623     \\
            RNN-RNN-8     & 0.1643     \\
            RNN-RNN-4     &  0.1666     \\
            RNN-ATT-RNN-2     & 0.1676   \\ 
            RVI          & 0.1751   \\
            RNN-ATT-RNN-4    & 0.1796   \\
            RNN-ATT-RNN-8    & 0.1903 \\
            \midrule
         \end{tabular}
         }
     \end{minipage}
     \begin{minipage}{0.16\textwidth}
      \resizebox{\linewidth}{!}{
    	\begin{tabular}{@{}c|c@{}}
        	\toprule
            Model   & \tiny{nDCG}@20 \\
            \midrule
            IMF           &  0.0523     \\
            CBoI-FF-1    & 0.1215    \\
            CBoI-FF-8    & 0.1229     \\
            CBoI-FF-4    &  0.1241    \\
 	        CBoI-FF-2    & 0.1242      \\
            RNN-FF-2    & 0.1242        \\
            RVI          & 0.1793   \\
            RNN-RNN-1     & 0.1833      \\
            RNN-FF-1    & 0.1839   \\
            Item-CF & 0.1867   \\
            RNN-FF-4    & 0.1875      \\
            RNN-RNN-2     &  0.1886      \\
            RNN-FF-8    & 0.1906     \\
            RNN-ATT-RNN-1     &  0.1913    \\
            RNN-RNN-8     & 0.1925     \\
            RNN-RNN-4     &  0.1943     \\
            RNN-ATT-RNN-2     & 0.1962    \\ 
            RNN-ATT-RNN-4    & 0.2054  \\
            RNN-ATT-RNN-8    & 0.2140 \\
            \midrule
         \end{tabular}
         }
     \end{minipage}
     
      \begin{minipage}{0.90\textwidth}
        \centering
        (b) RecSys15
    \end{minipage}  
    \caption{Recall, Precision and nDCG on MovieLens and RecSys15. Results are sorted by each metric.}\label{result}
\end{table*}

To explore the full potential of our model, there are several areas in which more research needs to be done. First, to better understand our model, more thorough analysis on the learned mixture components and the attention weights should be conducted. Second, we use embeddings pretrained using the word2vec objective, which leads to embeddings that learn the distributional, user-behavior based properties of items. One way to extend our model is to incorporate content-based attributes into the item embeddings we use, and create a hybrid recommender system.

\subsection*{Acknowledgments}
TW sincerely thanks Tommy Chen, Andrew Drozdov, Daniel Galron, Timothy Heath, Alex Shen, Krutika Shetty, Stephen Wu, Lijia Xie, and Kelly Zhang for helpful discussions and insightful feedbacks.	KC thanks support by eBay, TenCent, Facebook, Google and NVIDIA, and was partly supported by Samsung Advanced Institute of Technology (Next Generation Deep Learning: from pattern recognition to AI).

\bibliography{rec-mdn}
\bibliographystyle{aaai}

\end{document}